\documentclass[
]{ceurart}

\usepackage{listings}
\usepackage{graphicx}
\usepackage{amsmath}
\usepackage[ruled,vlined]{algorithm2e}  
\usepackage{float}
\usepackage{placeins}
\usepackage{placeins}

\begin{document}

\copyrightyear{2025}
\copyrightclause{Copyright for this paper by its authors.
  Use permitted under Creative Commons License Attribution 4.0
  International (CC BY 4.0).}

\conference{NORA'25: 1\textsuperscript{st} Workshop on Knowledge Graphs \& Agentic Systems Interplay co-located with NeurIPS, Dec.1, 2025, Mexico City, Mexico}

\title{Elastic Weight Consolidation for Knowledge Graph Continual Learning: An Empirical Evaluation}

\author[1]{Gaganpreet Jhajj}[%
orcid=0000-0001-5817-0297,
email=gjhajj1@learn.athabascau.ca,
]
\cormark[1]
\address[1]{School of Computing and Information Systems, Athabasca University, Canada}

\author[1]{Fuhua Lin}[%
orcid=0000-0002-5876-093X,
email=oscarl@athabascau.ca,
]

\begin{abstract}
Knowledge graphs (KGs) require continual updates as new information emerges, but neural embedding models suffer from catastrophic forgetting when learning new tasks sequentially. We evaluate Elastic Weight Consolidation (EWC), a regularization-based continual learning method, on KG link prediction using TransE embeddings on FB15k-237. Across multiple experiments with five random seeds, we find that EWC reduces catastrophic forgetting from 12.62\% to 6.85\%, a 45.7\% reduction compared to naive sequential training. We observe that the task partitioning strategy affects the magnitude of forgetting: relation-based partitioning (grouping triples by relation type) exhibits 9.8 percentage points higher forgetting than randomly partitioned tasks (12.62\% vs 2.81\%), suggesting that task construction influences evaluation outcomes. While focused on a single embedding model and dataset, our results demonstrate that EWC effectively mitigates catastrophic forgetting in KG continual learning and highlight the importance of evaluation protocol design.
\end{abstract}


\maketitle

\section{Introduction}

Knowledge graphs (KGs) depict structured information as networks of entities and their relations \citep{sheth2019knowledge}, facilitating a variety of applications, including question answering \citep{jhajj2024jack} and recommendation systems \citep{guo2020kgrec} and educational systems \citep{jhajj_educational_2024,kabir_llm-powered_2023,Lin2025,Jhajj_JPllmasr_2025,Gust2508cog,Gustafson2024}. These real-world KGs evolve continuously as new information becomes available and existing knowledge is refined. Neural embedding models, such as TransE \citep{bordes_translating_2013}, generate vector representations of entities and relations for link prediction. However, adapting these models to accommodate new information while retaining prior knowledge presents a significant challenge.

Catastrophic forgetting occurs when neural networks, trained sequentially on multiple tasks, experience significant performance degradation on earlier tasks after learning new ones \citep{mccloskey_catastrophic_1989}. This phenomenon poses particular challenges for KG embeddings, where maintaining consistent representations across evolving information is essential. While continual learning methods have been developed for image classification and natural language processing, their effectiveness on KG link prediction remains underexplored.

We investigate how Elastic Weight Consolidation (EWC) \citep{kirkpatrick_overcoming_2017}, a regularization-based continual learning method, performs on KG link prediction. EWC protects important parameters learned in previous tasks by adding a quadratic penalty based on the Fisher Information Matrix, allowing networks to learn new tasks while preserving performance on old ones. We evaluate EWC on TransE embeddings using FB15k-237, a standard KG benchmark, and compare relation-based task partitioning (where triples are grouped by relation type) to random partitioning.

Our experiments reveal that EWC substantially reduces catastrophic forgetting. On relation-based partitioned tasks, naive sequential training results in 12.62\% forgetting (measured as MRR degradation from post-task performance), while EWC with regularization strength $\lambda=10$ reduces this to 6.85\%, a 45.7\% reduction. This demonstrates that regularization-based continual learning effectively preserves KG embeddings across sequential tasks.

We also observe that the task partitioning strategy significantly affects the measured forgetting. Naive sequential training on relation-based tasks exhibits 12.62\% forgetting, compared to only 2.81\% on randomly partitioned tasks, a 9.8\%point difference. This suggests that evaluation protocols, particularly how tasks are constructed from datasets, influence continual learning metrics and should be carefully explored when it comes time to design the study.

Our study focuses on TransE embeddings for the FB15k-237 dataset across four relation-based tasks. While this scope limits generalizability, it provides rigorous evidence that EWC reduces catastrophic forgetting in KG continual learning and raises essential questions about task construction in continual learning evaluation.

This work addresses a critical challenge for knowledge graph-based AI agents: maintaining knowledge representations as new information arrives. As agents increasingly rely on KG-based memory and reasoning, the ability to incorporate new knowledge while preserving existing facts becomes essential for long-term autonomous operation.

\section{Related Work}

\textbf{KG Embeddings.} TransE \citep{bordes_translating_2013} represents relations as translations in embedding space, learning vectors such that $\mathbf{h} + \mathbf{r} \approx \mathbf{t}$ for true triples $(h,r,t)$. Extensions include TransH \citep{wang_knowledge_2014}, RotatE \citep{sun_rotate:_2019}, and ComplEx \citep{trouillon_complex_2016}.

\textbf{Continual Learning.} Methods for mitigating catastrophic forgetting include regularization approaches like EWC \citep{kirkpatrick_overcoming_2017} and Learning without Forgetting \citep{li_learning_2017}, replay-based methods that store and revisit previous examples \citep{rolnick_experience_2019}, and architectural approaches that allocate separate parameters for different tasks \citep{rusu_progressive_2022}. EWC estimates parameter importance using the Fisher Information Matrix and adds regularization penalties to protect important weights during subsequent task training.

\textbf{KG Continual Learning.} Prior work has explored continual learning for KGs across various contexts. \citet{wang_multi-task_2019} studied multi-task feature learning for KG-enhanced recommendation. \citet{Daruna2021} demonstrated continual learning methods for KG embeddings in robotic manipulation tasks, evaluating multiple architectures including TransE, DistMult, and ComplEx. \citet{Zhao2025} introduced the PS-CKGE benchmark, focusing on pattern shifts and demonstrating that these shifts exacerbate catastrophic forgetting beyond what would be expected from simple data scaling. Recent work has also examined embedding adaptation \citep{delange_continual_2021} and temporal KGs. While these comprehensive benchmarking efforts establish the landscape of KG continual learning, a focused empirical analysis of classic regularization methods like EWC, with explicit investigation of task partitioning effects, remains limited, motivating our study.

\section{Methodology}

\subsection{Problem Formulation}

A KG is $\mathcal{G} = (\mathcal{E}, \mathcal{R}, \mathcal{T})$ where $\mathcal{E}$ is the set of entities, $\mathcal{R}$ is the set of relations, and $\mathcal{T} \subseteq \mathcal{E} \times \mathcal{R} \times \mathcal{E}$ is the set of true triples. TransE learns embeddings $\mathbf{h}, \mathbf{r}, \mathbf{t} \in \mathbb{R}^d$ by minimizing:
\begin{equation}
\mathcal{L} = \sum_{(h,r,t) \in \mathcal{T}} \sum_{(h',r,t') \in \mathcal{T}'} \max(0, \gamma + d(\mathbf{h}+\mathbf{r},\mathbf{t}) - d(\mathbf{h}'+\mathbf{r},\mathbf{t}'))
\end{equation}
where $\mathcal{T}'$ contains negative samples, $d(\cdot,\cdot)$ is L2 distance, and $\gamma$ is the margin.

In continual learning, we partition $\mathcal{G}$ into tasks $\mathcal{G}_1, \ldots, \mathcal{G}_T$ and train sequentially. After training on task $i$, we measure performance $M_i^j$ on task $j \leq i$. Forgetting for task $j$ after learning task $i$ is:
\begin{equation}
F_i^j = M_j^j - M_i^j \quad \text{for } i > j
\end{equation}

We report average forgetting at the end of training:
\begin{equation}
\bar{F} = \frac{1}{T-1} \sum_{j=1}^{T-1} F_T^j
\end{equation}

\subsection{Elastic Weight Consolidation}

EWC protects important parameters by adding a regularization term to the loss when training on task $i$:
\begin{equation}
\mathcal{L}_{\text{EWC}}^i = \mathcal{L}^i + \frac{\lambda}{2} \sum_k F_k (\theta_k - \theta_{k,i-1}^*)^2
\end{equation}
where $\theta_{k,i-1}^*$ are optimal parameters after task $i-1$, $F_k$ is the Fisher Information diagonal approximation, and $\lambda$ controls regularization strength. The Fisher Information Matrix diagonal is:
\begin{equation}
F_k = \mathbb{E}_{(h,r,t) \sim \mathcal{G}_{i-1}} \left[ \left( \frac{\partial \log p(y|x;\theta)}{\partial \theta_k} \right)^2 \right]
\end{equation}

We compute this empirically using all triples from the previous task, processed in mini-batches (see Appendix \ref{app:A} for implementation details).

\subsection{Task Partitioning} 

We evaluate two task partitioning strategies to assess how task construction affects catastrophic forgetting: 

\textbf{Relation-based partitioning:} We partition FB15k-237 by grouping all triples that share the same relations. To create balanced task sizes, we sort the 237 relations by frequency (number of triples per relation) and assign them to four tasks in round-robin order: the most frequent relation goes to Task 1, the second-most to Task 2, the third-most to Task 3, the fourth-most to Task 4, the fifth-most back to Task 1, and so forth. This ensures each task receives approximately 59 relations with a mixture of common and rare relation types, while maintaining \textit{relation-level coherence}---all triples with the same relation appear in the same task. 

\textbf{Random partitioning:} We randomly shuffle all 272,115 training triples and divide them into four equal chunks of approximately 68,000 triples each. This distributes relation types across all tasks, creating \textit{relation-level overlap}---most relations appear in multiple tasks. The key difference: relation-based partitioning creates distinct distribution shifts between tasks (each task focuses on different relation types). In contrast, random partitioning minimizes distribution shift (each task is a representative sample of all relations). This allows us to isolate the effect of task boundary definition on the difficulty of continual learning.

\section{Experimental Setup}

\textbf{Dataset and Partitioning.} We use FB15k-237 \citep{toutanova_observed_2015}, which contains 14,505 entities, 237 relations, and 272,115 triples. Using relation-based partitioning (round-robin assignment by relation frequency), we create four tasks with approximately balanced sizes, each containing ~59 relations and their associated triples.

\textbf{Model Configuration.} We use TransE with 50-dimensional embeddings, margin $\gamma=1.0$, and L2 distance. Training uses the Adam optimizer \citep{kingma2017adammethodstochasticoptimization} with a learning rate of 0.001, a batch size of 256, and 20 epochs per task. We found 20 epochs sufficient for convergence on FB15k-237 in preliminary experiments.

\textbf{Methods Evaluated.} We compare naive sequential training (no continual learning) with EWC at multiple regularization strengths ($\lambda \in \{0.1, 1.0, 10.0\}$), EWC combined with experience replay (500 examples per task), and replay-only baselines (random and wave-based sampling).

\textbf{Evaluation Protocol.} We use Mean Reciprocal Rank (MRR) for link prediction, computing filtered rankings that exclude known actual triples. For each task $j$, we record MRR immediately after training ($M_j^j$) and after each subsequent task ($M_i^j$ for $i > j$). We run five random seeds (42, 123, 456, 789, 2024) and report means and standard deviations.

\textbf{Hardware.} Experiments ran on NVIDIA RTX 3070 Ti (8GB) with ~20 hours total computation for 80 experiments (8 methods $\times$ 5 seeds $\times$ 2 partitioning strategies).

\section{Results}

\subsection{Classical Methods on Relation-Based Partitioning}

Table \ref{tab:semantic_results} shows forgetting results for classical continual learning methods on relation-based partitioned tasks.

\begin{table}[h]
\centering
\small
\begin{tabular}{lcc}
\toprule
\textbf{Method} & \textbf{Forgetting (\%)} & \textbf{Final MRR} \\
\midrule
Naive & 12.62 $\pm$ 0.35 & 0.206 $\pm$ 0.006 \\
EWC ($\lambda=0.1$) & 10.44 $\pm$ 0.26 & 0.229 $\pm$ 0.005 \\
EWC ($\lambda=1.0$) & 7.51 $\pm$ 0.44 & 0.250 $\pm$ 0.006 \\
\textbf{EWC ($\lambda=10$)} & \textbf{6.85 $\pm$ 0.33} & \textbf{0.242 $\pm$ 0.004} \\
EWC + Wave Replay & 9.91 $\pm$ 0.20 & 0.234 $\pm$ 0.005 \\
Random Replay & 13.78 $\pm$ 0.44 & 0.196 $\pm$ 0.006 \\
Wave Replay & 12.54 $\pm$ 0.14 & 0.216 $\pm$ 0.007 \\
\bottomrule
\end{tabular}
\caption{Forgetting and final MRR on relation-based partitioned tasks (mean $\pm$ std over 5 seeds).}
\label{tab:semantic_results}
\end{table}

EWC with $\lambda=10$ achieves the best performance, reducing forgetting to 6.85\% (std 0.33\%), a 45.7\% reduction compared to naive training. This demonstrates that regularization-based protection of important parameters effectively mitigates catastrophic forgetting in KG continual learning. Final MRR also improves from 0.206 to 0.242, indicating that EWC preserves not only previous task performance but also maintains overall embedding quality.

Interestingly, replay-based methods underperform. Random replay achieves 13.78\% forgetting, worse than naive training, suggesting that simply revisiting old examples without principled parameter protection may interfere with learning. Wave-based replay performs similarly (12.54\% forgetting). Combining EWC with wave replay (9.91\% forgetting) improves over replay alone but underperforms pure EWC, indicating that regularization is the primary driver of performance.

\begin{figure}[h]
\centering
\includegraphics[width=0.9\textwidth]{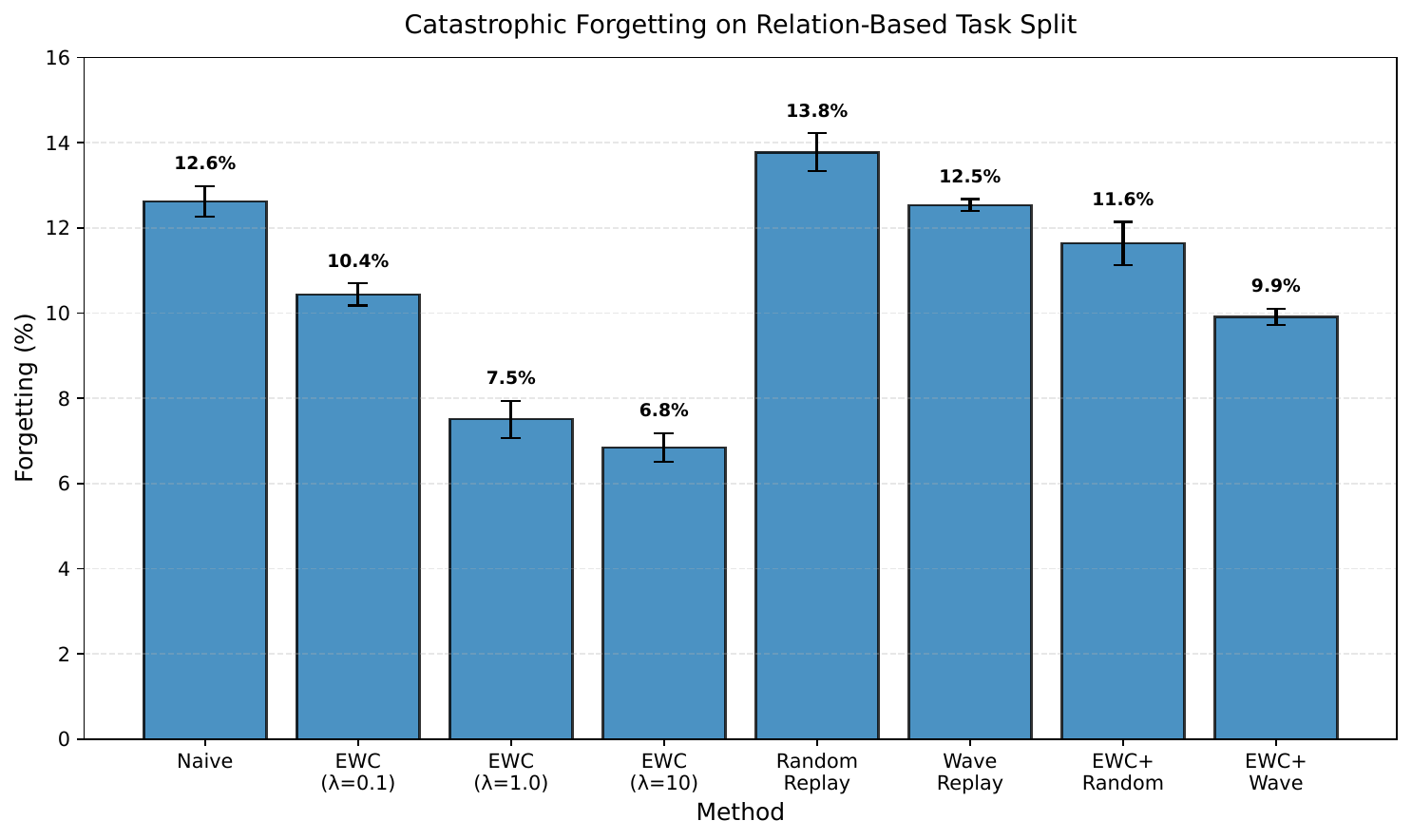}
\caption{Catastrophic forgetting on relation-based partitioned tasks. EWC ($\lambda=10$) substantially reduces forgetting compared to naive sequential training and replay-based methods.}
\label{fig:semantic}
\end{figure}

Figure \ref{fig:semantic} visualizes these results, showing a clear separation between EWC and other methods.

\subsection{Effect of Task Partitioning}

We compare relation-based and random partitioning strategies to assess whether task construction affects measured forgetting. Table \ref{tab:partitioning} and Figure \ref{fig:partitioning} show results.

\begin{table}[h]
\centering
\small
\begin{tabular}{lcc}
\toprule
\textbf{Partitioning} & \textbf{Naive Forgetting (\%)} & \textbf{EWC Forgetting (\%)} \\
\midrule
Relation-based & 12.62 $\pm$ 0.35 & 6.85 $\pm$ 0.33 \\
Random & 2.81 $\pm$ 0.34 & 5.08 $\pm$ 0.22 \\
\textbf{Difference} & \textbf{9.81 pp} & \textbf{1.77 pp} \\
\bottomrule
\end{tabular}
\caption{Forgetting comparison between relation-based and random task partitioning (mean $\pm$ std over 5 seeds).}
\label{tab:partitioning}
\end{table}

\begin{figure}[h]
\centering
\includegraphics[width=0.9\textwidth]{semantic_vs_random_new.pdf}
\caption{Effect of task partitioning on forgetting. Relation-based partitioning creates more challenging continual learning scenarios, with 9.8 percentage points of higher forgetting in naive training compared to random partitioning.}
\label{fig:partitioning}
\end{figure}

Relation-based partitioning results in substantially higher forgetting during naive training (12.62\% vs 2.81\%), a 9.8 percentage-point difference. We hypothesize this occurs because relation-based partitioning creates \textit{task coherence}: each task focuses on a distinct subset of relation types (approximately 59 relations per task), inducing larger distribution shifts when transitioning between tasks. When the model moves from Task 1's relations to Task 2's completely different relation set, the parameter updates required are more disruptive to previously learned representations. In contrast, random partitioning creates \textit{relation-level overlap}: each task contains a representative sample of most relation types, so the distribution shift between tasks is minimal. This naturally regularizes learning, as the model continually encounters all relation types across tasks. This observation suggests that task construction significantly influences the difficulty of continual learning and that evaluation protocols should explicitly consider and report partitioning strategies.

Notably, EWC reduces this gap: the difference between relation-based and random partitioning under EWC is only 1.77 percentage points (6.85\% vs 5.08\%), suggesting that effective continual learning methods can generalize across different task construction approaches.

\section{Discussion}

KG embeddings have structured parameter spaces where specific dimensions encode semantic properties. The Fisher Information Matrix identifies parameters critical for encoding relation types and entity characteristics learned in previous tasks. By protecting these parameters, EWC enables new-task learning while preserving the semantic structure of the embedding space.

The superior performance of EWC compared to replay methods suggests that principled parameter protection is more effective than simply revisiting old examples when working memory (replay buffer) is limited. This aligns with neuroscience findings that synaptic consolidation, rather than replay alone, enables long-term memory retention.

Our EWC forgetting rate (6.85\%) on relation-based partitioned tasks demonstrates effective continual learning for KG link prediction. While direct comparison with prior work is challenging due to differences in datasets and evaluation protocols, our results are consistent with EWC's performance on image classification tasks \citep{kirkpatrick_overcoming_2017} and suggest that regularization-based continual learning generalizes to structured knowledge representations.

The task partitioning effect we observe (a 9.8 percentage-point difference) highlights an important consideration for continual learning evaluation: reported forgetting rates depend on how tasks are constructed. This suggests that future work should explicitly consider task construction methodology when designing experiments and reporting results.

\section{Limitations and Future Work}

Our study has several limitations that define the scope and future directions. 

We evaluate only TransE on FB15k-237 across four tasks, a deliberate choice given consumer-grade GPU constraints (NVIDIA RTX 3070 Ti, 8GB) that enabled rigorous multi-seed experimentation within feasible time scales. Results may not generalize to: (1) complex embedding methods (RotatE, ComplEx, TuckER); (2) other datasets (WN18RR, YAGO, Wikidata); (3) more extended task sequences (10+ tasks). We compare EWC with basic replay baselines, but do not exhaustively benchmark all continual learning methods. Our relation-based partitioning uses round-robin frequency assignment; alternative strategies (entity-based, domain-based) warrant investigation. 

Experiments were constrained to academic resources without HPC clusters or large-scale GPU infrastructure. While this motivated our focused design, a comprehensive multi-model evaluation would require substantially more resources. We view this as proof-of-concept evidence for future large-scale studies.

Future work should evaluate EWC across multiple embedding methods and datasets to assess generalizability. Scaling studies with 10+ tasks would reveal the long-term dynamics of continual learning. Systematic investigation of task construction strategies could formalize the relationship between task partitioning and forgetting. Additionally, combining EWC with more sophisticated replay strategies or architectural approaches may yield further improvements.

A particularly promising direction is to evaluate continual learning methods on educational KGs, where knowledge evolves as curriculum content updates and student learning data accumulates. Recent work has highlighted the potential of open-source LLMs to transform educational contexts through adaptive and personalized learning \cite{Lin2024}. However, the challenge of maintaining such systems as they continually learn from new educational content remains underexplored. Building on recent work in curriculum modeling and adaptive learning \citep{Lin2025}, we plan to investigate how EWC performs when domain-specific educational KGs are continually updated with new learning resources, pedagogical relations, and student interaction patterns. The integration of open-source LLMs (e.g., Llama \cite{touvron2023llamaopenefficientfoundation}, Mistral \cite{jiang2023mistral7b}, Qwen \cite{qwen2025qwen25technicalreport}) with educational KGs for personalized learning support presents unique continual learning challenges: both the symbolic KG structure and LLM parameters must adapt to new content while preserving existing pedagogical knowledge. Our findings on task partitioning effects may inform how educational content updates should be structured to minimize interference with previously learned material.

Neuromorphic approaches using spiking neural networks with spike-timing-dependent plasticity offer promising directions for extending this work. Building on prior work demonstrating SNN-based relational inference and knowledge representation for knowledge graphs \citep{jhajj_neuromorphic_2025}, we plan to investigate whether the biological learning mechanisms inherent in STDP can provide natural solutions to catastrophic forgetting. The structured, relational nature of KGs may align particularly well with neuromorphic computation, where synaptic plasticity mechanisms could enable task-consolidation without explicit regularization.

While our preliminary experiments on consumer-grade GPUs were inconclusive, a comprehensive evaluation using specialized neuromorphic hardware (Intel Loihi, IBM TrueNorth, SpiNNaker) could reveal whether biological learning mechanisms provide advantages for continual learning in knowledge graphs. The structured, relational nature of KGs may align well with neuromorphic computation, particularly for consolidating knowledge across tasks. Given EWC's success through parameter-wise importance weighting, bio-inspired plasticity mechanisms that selectively strengthen or weaken synaptic connections based on usage patterns warrant thorough exploration with appropriate hardware infrastructure.

\section{Conclusion}

We evaluated EWC for KG continual learning using TransE embeddings on FB15k-237. Across multiple experiments with five random seeds, we found that EWC reduces catastrophic forgetting from 12.62\% to 6.85\%, a 45.7\% reduction compared to naive sequential training. This demonstrates that regularization-based continual learning effectively preserves KG embeddings across sequential tasks.

We also observed that the task partitioning strategy significantly affects measured forgetting: relation-based partitioning exhibits 9.8 percentage points higher forgetting than randomly partitioned tasks for naive training. This suggests that the evaluation protocol's design, particularly the task-construction methodology, influences continual learning measurements and should be carefully considered in experimental design.

While our study focuses on a single embedding model and dataset, it provides rigorous evidence for EWC's effectiveness and raises essential questions about evaluation methodology in KG continual learning.
\begin{acknowledgments}

We acknowledge the support of the Natural Sciences and Engineering Research Council of Canada (NSERC), Alberta Innovates, Alberta Advanced Education, and Athabasca University, Canada. We would also like to thank the reviewers for their suggestions on how to improve this work.
\end{acknowledgments}


\section*{Declaration on Generative AI}

During the preparation of this work, the author(s) used Grammarly and Claude (Anthropic) for Grammar and spelling checks.

\newpage

\bibliography{sample-ceur}


\clearpage
\appendix

\section{EWC Implementation Details}
\label{app:A}

We compute the Fisher Information Matrix diagonal after training each task using all triples from that task, processed in mini-batches of 256. For each batch $b$, we compute the empirical Fisher approximation:

\begin{equation}
F_k \approx \frac{1}{B} \sum_{b=1}^{B} \left( \frac{\partial \mathcal{L}_b}{\partial \theta_k} \right)^2
\end{equation}

where $B$ is the total number of batches in the task and $\mathcal{L}_b$ is the loss on batch $b$. This accumulates squared gradients across all training examples from the previous task.

During training on task $i$, we apply the EWC penalty for all previous tasks:
\begin{equation}
\mathcal{L}_{\text{total}}^i = \mathcal{L}^i + \sum_{j=1}^{i-1} \frac{\lambda}{2} \sum_k F_k^j (\theta_k - \theta_{k,j}^*)^2
\end{equation}

\section{Random Partitioning Results}
\label{app:B}
Table \ref{tab:random_full} shows full results for random partitioning.

\begin{table}[h]
\centering
\small
\begin{tabular}{lcc}
\toprule
\textbf{Method} & \textbf{Forgetting (\%)} & \textbf{Final MRR} \\
\midrule
Naive & 2.81 $\pm$ 0.34 & 0.277 $\pm$ 0.008 \\
EWC ($\lambda=0.1$) & 2.88 $\pm$ 0.34 & 0.270 $\pm$ 0.005 \\
EWC ($\lambda=1.0$) & 3.88 $\pm$ 0.27 & 0.244 $\pm$ 0.005 \\
EWC ($\lambda=10$) & 5.08 $\pm$ 0.22 & 0.222 $\pm$ 0.008 \\
\bottomrule
\end{tabular}
\caption{Full results on random partitioning.}
\label{tab:random_full}
\end{table}

Interestingly, stronger EWC regularization increases forgetting on random partitioning (5.08\% for $\lambda=10$ vs 2.81\% naive). We hypothesize that random partitioning naturally distributes relation types across tasks, reducing interference. Strong regularization may over-constrain parameters, preventing necessary adaptation. This suggests that optimal regularization strength depends on task construction.

\section{Hyperparameter Sensitivity}
\label{app:C}
Table \ref{tab:lambda_sweep} shows EWC performance across $\lambda$ values we tested.

\begin{table}[h]
\centering
\small
\begin{tabular}{lccc}
\toprule
\textbf{$\lambda$} & \textbf{Relation-based (\%)} & \textbf{Random (\%)} & \textbf{Final MRR (Relation-based)} \\
\midrule
0.1 & 10.44 $\pm$ 0.26 & 2.88 $\pm$ 0.34 & 0.229 $\pm$ 0.005 \\
1.0 & 7.51 $\pm$ 0.44 & 3.88 $\pm$ 0.27 & 0.250 $\pm$ 0.006 \\
10.0 & 6.85 $\pm$ 0.33 & 5.08 $\pm$ 0.22 & 0.242 $\pm$ 0.004 \\
\bottomrule
\end{tabular}
\caption{EWC performance across tested regularization strengths.}
\label{tab:lambda_sweep}
\end{table}

For relation-based partitioning, $\lambda=10$ achieves the lowest forgetting. For random partitioning, weaker regularization ($\lambda=0.1$) performs best. This suggests that optimal regularization strength depends on task construction: relation-grouped tasks require stronger protection of essential parameters, while randomly distributed tasks benefit from more flexibility.
\newpage

\section{Performance-Forgetting Trade-off}
\label{app:D}

\begin{figure}[h!]
\centering
\includegraphics[width=0.8\textwidth]{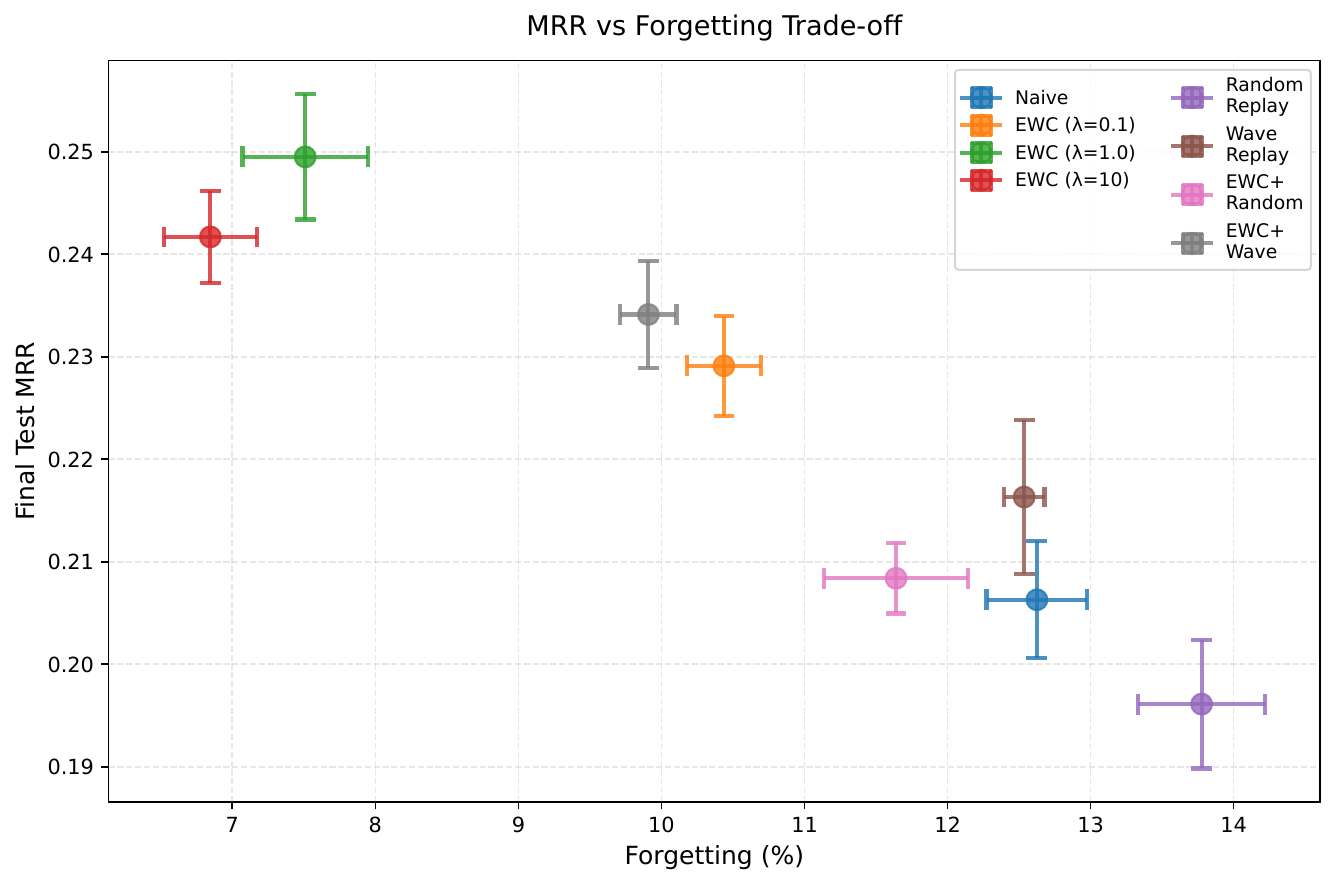}
\caption{Performance-forgetting trade-off. EWC ($\lambda=10$) achieves the best balance, with low forgetting (6.85\%) and competitive final MRR (0.242).}
\label{fig:tradeoff}
\end{figure}

Figure \ref{fig:tradeoff} visualizes the relationship between final MRR and forgetting for all methods.

\FloatBarrier
\FloatBarrier

EWC with $\lambda=10$ occupies the optimal region (low forgetting, competitive performance). Replay methods cluster in the high-forgetting, low-performance region. This visualization confirms that EWC achieves superior trade-offs compared to alternative approaches.

\section{Implementation Details}
\label{app:F}

\textbf{Code Structure.} Experiments implemented in PyTorch 1.13. 

\textbf{Negative Sampling.} We use uniform random tail corruption with a 1:1 ratio of positive to negative samples. For each positive triple $(h,r,t)$, we generate one negative sample $(h,r,t')$ by randomly replacing the tail entity $t'$ sampled uniformly from $\mathcal{E}$. We use unfiltered negative sampling during training (negative samples may coincidentally be true triples), but employ filtered evaluation for link prediction metrics.

\textbf{Hardware.} Experiments ran on NVIDIA RTX 3070 Ti (8GB) with approximately 18-20 hours total computation for all experimental runs (8 methods $\times$ 5 seeds $\times$ 2 partitioning strategies). 

\newpage

\section{Task Retention Visualization}
\label{app:E}

\begin{figure}[h]
\centering
\includegraphics[width=0.75\textwidth]{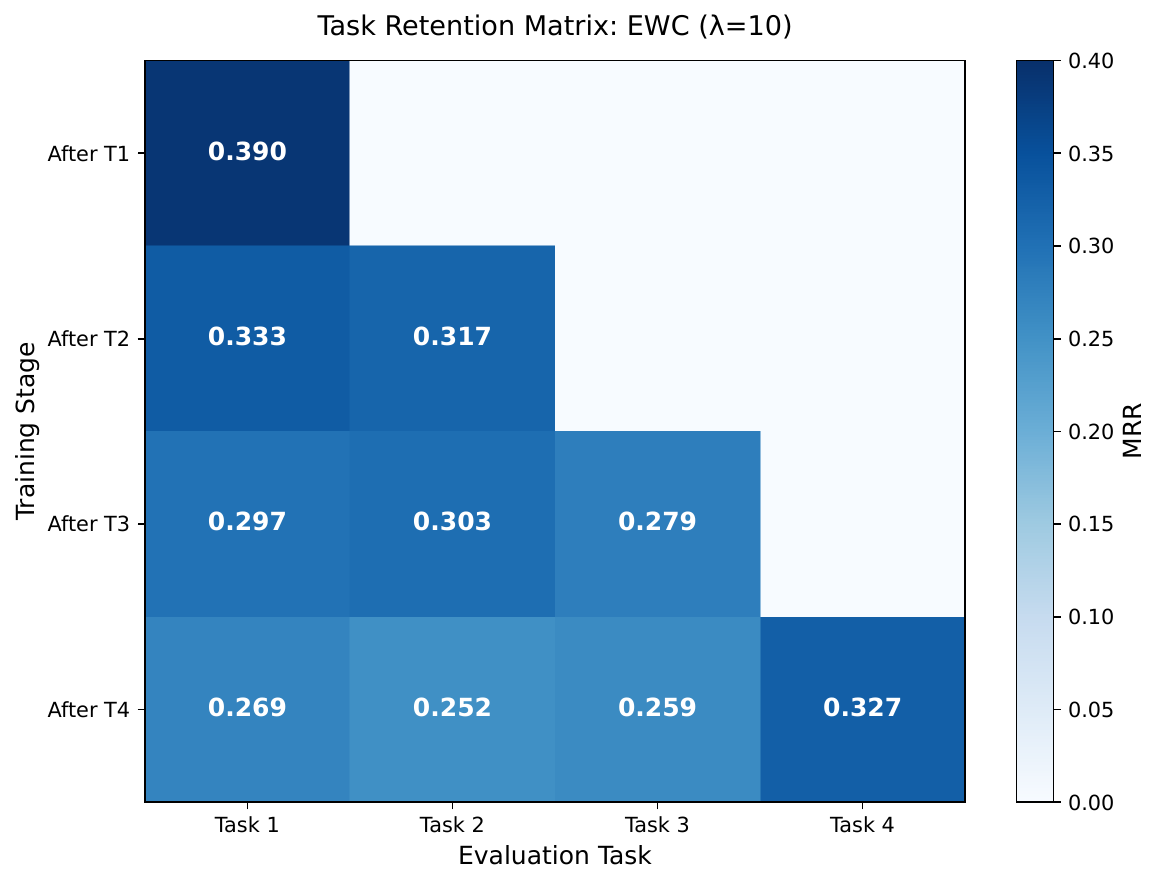}
\caption{Task retention matrix for EWC ($\lambda=10$). The diagonal shows performance immediately after learning each task; the line below the diagonal shows retention after subsequent tasks. Minimal degradation indicates effective continual learning.}
\label{fig:retention}
\end{figure}

Figure \ref{fig:retention} shows how performance on each task degrades over subsequent training.

The heatmap shows that EWC maintains relatively stable performance across tasks, with limited degradation on earlier tasks as new tasks are learned. This visualization confirms that EWC effectively protects previous task performance.

\end{document}